\documentclass{article}

\usepackage{PRIMEarxiv}

\usepackage[utf8]{inputenc} 
\usepackage[russian]{babel}
\usepackage[T2A]{fontenc}    
\usepackage{hyperref}       
\usepackage{url}            
\usepackage{booktabs}       
\usepackage{amsfonts}       
\usepackage{nicefrac}       
\usepackage{microtype}      
\usepackage{lipsum}
\usepackage{fancyhdr}       
\usepackage{graphicx}       
\graphicspath{{media/}}     

\title{IVR на базе искусственного интеллекта для задач колл-центров
}

\author{
  Кошербай Гасырбек \\
  Softlance LLP\\
  г. Уральск\\
  \texttt{gassyrbek2001@gmail.com} \\
   \And
  Нургиса Апбаз \\
  Softlance LLP \\
  г. Уральск\\
  \texttt{apbaz.nurgisa@gmail.com} \\
}

\makeatletter
\renewenvironment{abstract}{
    \if@twocolumn
      \section*{\abstractname}%
    \else
      \begin{center}%
        {\bfseries \large\abstractname\vspace{-.5em}\vspace{\z@}}%
      \end{center}%
      \quote
    \fi}
    {\if@twocolumn\else\endquote\fi}
\makeatother

\begin{document}
\maketitle

\begin{abstract}
Использование традиционных методов IVR (Interactive Voice Response) часто оказывается недостаточным для удовлетворения потребностей клиентов. В данной статье рассматривается применение технологий искусственного интеллекта (ИИ) для повышения эффективности работы IVR-систем в колл-центрах.
Предлагается подход, основанный на интеграции решений для преобразования речи в текст, классификации текстовых запросов с помощью больших языковых моделей (LLM) и синтеза речи. Особое внимание уделяется адаптации этих технологий к работе с казахским языком, включая Fine-Tuning моделей на специализированных датасетах. Описываются практические аспекты внедрения разработанной системы в реальный колл-центр классификации обращений.
Результаты исследования демонстрируют, что применение ИИ-технологий в IVR-системах колл-центров позволяет снизить нагрузку на операторов, повысить качество обслуживания клиентов и увеличить эффективность обработки обращений. Предложенный подход может быть адаптирован для использования в колл-центрах, работающих с различными языками.
\end{abstract}


\section{Введение}
Классические IVR-системы (Interactive Voice Response) в колл-центрах, основанные на голосовых меню и скриптах операторов, часто оказываются недостаточными для удовлетворения растущих потребностей современных клиентов. Клиенты ожидают более персонализированного и эффективного обслуживания, в то время как операторы сталкиваются с высокой нагрузкой при обработке большого количества разнообразных обращений.
В последние годы искусственный интеллект (ИИ) стал важным инструментом для решения многих задач, включая автоматизацию и оптимизацию процессов в колл-центрах. ИИ способен анализировать большие объемы данных, выявлять скрытые закономерности и предлагать решения, которые ранее были недоступны. Внедрение ИИ в IVR системы открывает новые возможности для повышения эффективности работы колл-центров, а конкретно решения для задач по преобразования речи в текст, классификации текстовых запросов и синтеза речи. Такие технологии позволят автоматизировать обработку обращений, снизить нагрузку на операторов и улучшить качество обслуживания клиентов.
Данная статья посвящена разработке и внедрению ИИ-решений в IVR-систему колл-центра, специализирующегося на классификации обращений. Мы рассмотрим особенности адаптации этих технологий к работе с казахским языком и также остановимся на практических аспектах реализации предложенного подхода.

\newpage
\section{Постановка задачи}

Современные колл-центры сталкиваются с рядом проблем, которые требуют внедрения передовых технологий для повышения эффективности и качества обслуживания клиентов. Основные проблемы, на решение которых направлено данное исследование, включают:
\begin{itemize}
    \item \textbf{Ограниченные возможности классических IVR-систем}
    
    Традиционные IVR-системы, основанные на голосовых меню и скриптах операторов, часто не справляются с растущими потребностями клиентов. Они не обеспечивают должного уровня персонализации и эффективности обработки разнообразных обращений.
    \item \textbf{Высокая нагрузка на операторов}

    Операторы колл-центров вынуждены обрабатывать большое количество разнообразных запросов, что приводит к высокой нагрузке и снижению качества обслуживания. Необходимы решения, способные автоматизировать обработку простых обращений.

    \item \textbf{Сложность классификации обращений}

    Классификация обращений клиентов является ключевой задачей для эффективной маршрутизации и дальнейшей обработки. Однако, ручная классификация операторами становится все более трудоемкой из-за растущего разнообразия запросов.

    \item \textbf{Адаптация к особенностям языка}

    Многие колл-центры работают с клиентами, говорящими на различных языках, включая языки с нетривиальными особенностями. В нашем случая это казахский язык. Необходимо обеспечить высокую точность распознавания речи и классификации запросов для казахского в том числе.
\end{itemize}

Для решения этих проблем данное исследование ставит следующие задачи

\subsection{Распознавание речи}
Первой и одной из наиболее важных задач является обработка голоса в реальном времени. Колл-центры должны обеспечивать быстрое и точное распознавание речи клиентов, что требует использования современных инструментов и технологий для обработки голоса. Это включает в себя использование систем автоматического распознавания речи (ASR), которые могут эффективно работать с разнообразными акцентами, интонациями и фоновыми шумами, но даже в наиболее современных открых решениях имеется недостаток, а именно работа с казахским языком. В дальнейшем мы подробно рассмотрим как решить эту проблему. В итоге мы имеем, что на входе в систему поступает голосовой запрос клиента, который преобразуется в текст с помощью ASR. И мы плавно перетекаем к следующей задаче

\subsection{Обработка текста}

Задача связана с последующей обработкой текста, полученного после распознавания речи. Необходимо реализовать эффективную классификацию текстовых запросов с помощью технологий искусственного интеллекта для определения проблемы клиента и дальнейшей маршрутизации обращения. Небольшое оступление - как упоминалось в главе 1, я указывал на проблему современных IVR в купе с большим объёмом данных. В нашем контексте - большим количеством классов. Клиент не обязан знать классификацию своей проблемы и точно её диктовать, также невозможно воспроизвести классический метод нажатия кнопок, так как число классов около 200.
Возвращаясь обратно, мы должны "валидировать" наш классификатор у клиента, чтобы в случая ошибки перенаправить на оператора.

\subsection{Преобразование текста в речь}

Необходимо интегрировать технологии синтеза речи для предоставления клиентам персонализированных ответов и улучшения взаимодействия. Подробнее про методы преобразования текста в речь мы рассмотрим в следующей главе.

\newpage
\section{Методология}
В данной главе мы рассмотрим детально каждую из задач и подходы к решению. Начнем с технологий распознавания речи. 
Автоматическое распознавание речи (Automatic Speech Recognition, ASR) - это технология, которая преобразует устную речь в текстовый формат \cite{graves2014towards} . Исследователи в области ASR, как правило, используют два основных подхода для обучения систем ASR: полностью контролируемые (fully supervised) или самообучающиеся (self-supervised) модели \cite{radford2021learning}.
В данной работе мы фокусируемся на использовании fully-supervised моделей, в частности, на модели Whisper, разработанной компанией OpenAI \cite{kaneko2017acoustic}.
\subsection{Whisper}
Модель Whisper является крупной языковой моделью, обученной на огромном количестве транскрибированных аудиоданных, что позволяет ей выполнять точное преобразование речи в текст на широком спектре языков \cite{kaneko2017acoustic}. Однако, когда мы работаем с языками, которые недостаточно представлены в исходных данных для обучения Whisper, таких как казахский язык, производительность модели может страдать, приводя к высоким показателям ошибки распознавания (Word Error Rate, WER) около 50 \% \cite{aitzhanov2019kazakh}
Для решения этой проблемы мы предлагаем провести дополнительную настройку (fine-tuning) модели Whisper на специализированном наборе данных казахской речи, предоставленном Казахским национальным университетом имени Аль-Фараби, который содержит 558 часов записей на казахском языке. Применяя методы fine-tuning, мы можем адаптировать модель Whisper к особенностям казахского языка и улучшить ее точность для этого конкретного случая использования.
Один из эффективных методов дообучения, который мы применяем, - это техника Low-Rank Adaptation (LoRA) \cite{hu2022lora}. LoRA позволяет эффективно адаптировать модель без значительного увеличения ее размера или вычислительных требований, что делает ее подходящим выбором для реального развертывания в среде колл-центра.
\begin{figure}[h]
    \centering
    \includegraphics[width=0.8\textwidth]{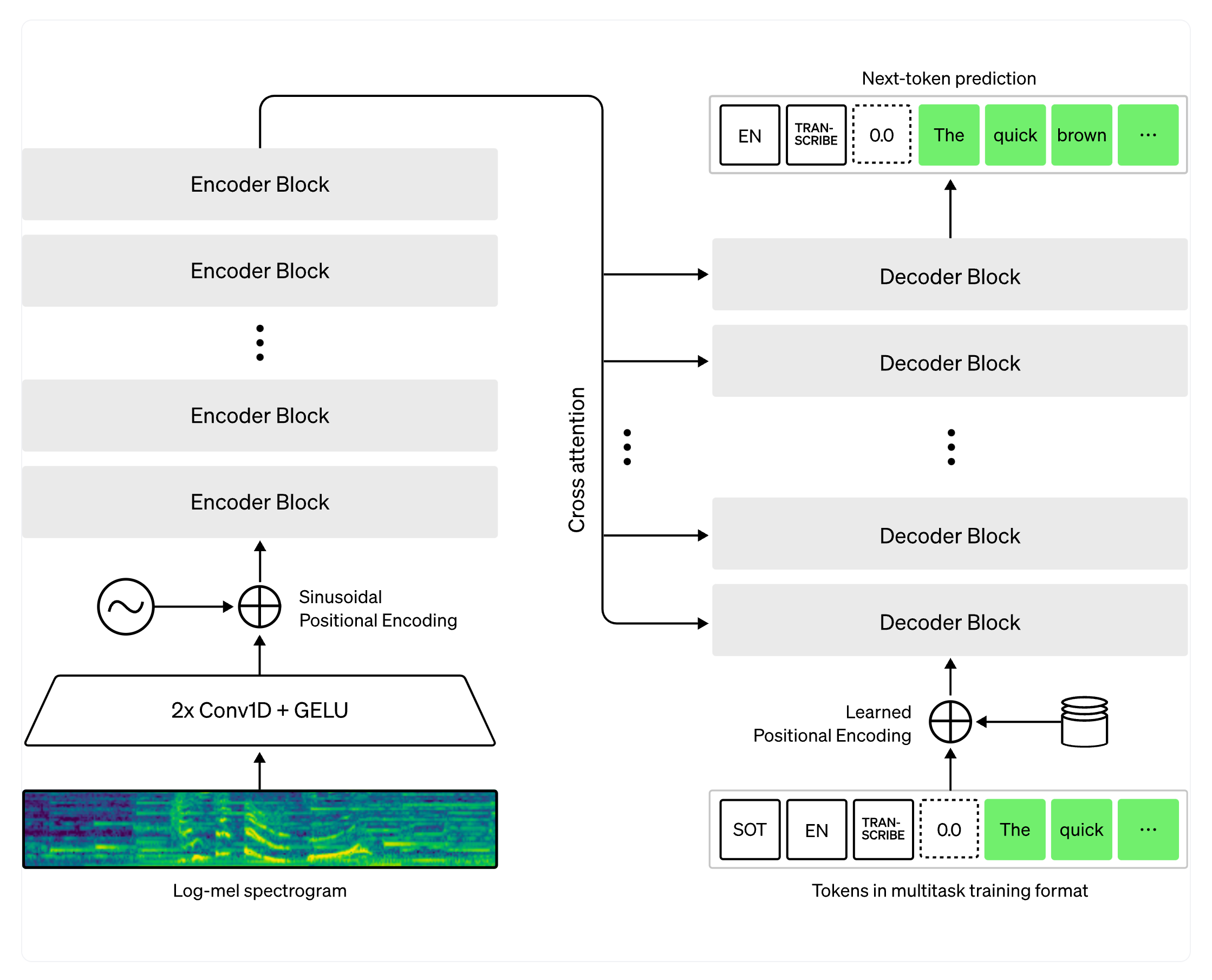}
    \caption{Архитектура модели Whisper}
    \label{fig:whisper_architecture}
\end{figure}
Используя модель Whisper и применяя к ней fine-tuning с помощью LoRA на наборе данных казахской речи, мы стремимся разработать высокоточную систему преобразования речи в текст, которая может справляться с уникальными особенностями казахского языка в контексте IVR-системы колл-центра.

\subsection{Large Language Models}
Помимо преобразования речи в текст, мы также используем большие языковые модели (Large Language Models, LLM) для классификации текстовых запросов клиентов и построения диалога. LLM - это мощные модели глубокого обучения, обученные на огромных объемах текстовых данных, которые способны эффективно выполнять широкий спектр языковых задач \cite{brown2020language}.
Для решения задачи классификации проблем клиентов мы выбрали модель IrbisGPT от компании MOST Holdings, которая обладает знаниями о казахском языке. Кроме того, мы также интегрируем технологию RAG (Retrieval-Augmented Generation) для обогащения модели дополнительной информацией.

Для эффективного использования больших языковых моделей, мы будем использовать Ollama. Ollama — это проект с открытым исходным кодом, который служит мощной и удобной платформой для запуска LLM на локальном компьютере, а также является безопасной с точки зрения конфиденциальности. Для запуска своей модели в Ollama необходимо заранее провести квантование - значительно уменьшение размера модели и снизижения требования к вычислительным ресурсам, не сильно теряя при этом в качестве предсказаний.

Все это дает нам возможность развертывать и использовать LLM в реальных условиях колл-центра, обеспечивая быстрое и эффективное предоставление ответов на запросы клиентов.

Важно, чтобы LLM модель была обучена на качественных датасетах, отражающих реальные сценарии взаимодействия с клиентами в колл-центре. Это позволит точно идентифицировать широкий спектр запросов.

Для дальнейшего улучшения диалога мы также интегрируем технологии преобразования текста в речь (Text-to-Speech), такие как mms-tts-kaz от META \cite{li2023scaling}.

Полный процесс можно описать следующим образом:

\begin{itemize}
    \item Поступающие речевые обращения клиентов преобразуются в текст с помощью модели Whisper.
    \item Текстовые запросы классифицируются с использованием IrbisGPT.
    \item Результаты классификации передаются в систему маршрутизации для дальнейшей обработки.
    \item Преобразования текста в речь (mms-tts-kaz) озвучивает классификацию обратно клиенту для валидации и уточнения.
    \item В случае негативной реакции со стороны клиента или неуверенности системы, запрос переводится на оператора для ручной обработки.
\end{itemize}

\section{Практическая часть и заключение}

Для преобразования речевых обращений клиентов в текст мы использовали модель Whisper-small. Проведя дообучение этой модели на датасете казахской речи (KSD), предоставленном Аль-Фарабийским университетом, мы добились снижения показателя ошибки распознавания (Word Error Rate, WER) до 16 \% на тестовых данных (не целевых со стороны колл-центра). Это значительно превосходит результаты, которые можно было бы получить без дополнительной настройки модели.
Дообучение проводилось на T4 GPU. Колл-центр построен на платформе Asterisk. Мы организовали Pipeline от распознания до синтеза речи.

В заключении можно сказать, что внедрение технологий искусственного интеллекта в IVR-системы колл-центров открывает широкие возможности для повышения эффективности и качества обслуживания клиентов. Разработанный нами подход, основанный на интеграции решений для преобразования речи в текст, классификации текстовых запросов и синтеза речи, продемонстрировал значительные практические результаты.

Дальнейшие исследования в этом направлении могут быть связаны с расширением функциональности ИИ-системы, повышением точности классификации запросов и интеграцией дополнительных модулей для улучшения взаимодействия с клиентами.


\newpage
\begin{thebibliography}{99}

\bibitem{kour2014real}
George Kour and Raid Saabne.
\textit{Real-time segmentation of on-line handwritten Arabic script}.
In Proceedings of the 14th International Conference on Frontiers in Handwriting Recognition (ICFHR), pages 417--422, IEEE, 2014.

\bibitem{kour2014fast}
George Kour and Raid Saabne.
\textit{Fast classification of handwritten on-line Arabic characters}.
In Proceedings of the 6th International Conference on Soft Computing and Pattern Recognition (SoCPaR), pages 312--318, IEEE, 2014.

\bibitem{graves2014towards}
Alex Graves and Navdeep Jaitly.
\textit{Towards end-to-end speech recognition with recurrent neural networks}.
In Proceedings of the International Conference on Machine Learning, pages 1764--1772, PMLR, 2014.

\bibitem{radford2021learning}
Alec Radford, Jong Wook Kim, Chris Hallacy, Aditya Ramesh, Gabriel Goh, Sandhini Agarwal, Girish Sastry, Amanda Askell, Pamela Mishkin, Jack Clark, et al.
\textit{Learning Transferable Visual Models From Natural Language Supervision}.
In Proceedings of the IEEE/CVF Conference on Computer Vision and Pattern Recognition, pages 8748--8758, 2021.

\bibitem{kaneko2017acoustic}
Eiji Kaneko, Yuya Fujii, and Chiori Hori.
\textit{Acoustic model adaptation using teacher-student training for end-to-end speech recognition}.
In Proceedings of the 2017 IEEE Automatic Speech Recognition and Understanding Workshop (ASRU), pages 163--170, IEEE, 2017.

\bibitem{aitzhanov2019kazakh}
A. Aitzhanov, T. Tolegenov, A. Sharipbay, and K. Alimhan.
\textit{Kazakh speech recognition using deep learning}.
In Proceedings of the 2019 IEEE 13th International Conference on Application of Information and Communication Technologies (AICT), pages 1--5, IEEE, 2019.

\bibitem{hu2022lora}
Edward J. Hu, Yelong Shen, Phillip Wallis, Zeyuan Allen-Zhu, Yuanzhi Li, Shean Wang, and Lu Wang.
\textit{LoRA: Low-Rank Adaptation of Large Language Models}.
In Proceedings of the International Conference on Learning Representations, 2022.

\bibitem{brown2020language}
Tom B. Brown, Benjamin Mann, Nick Ryder, Melanie Subbiah, Jared D. Kaplan, Prafulla Dhariwal, Arvind Neelakantan, Pranav Shyam, Girish Sastry, Amanda Askell, et al.
\textit{Language models are few-shot learners}.
Advances in Neural Information Processing Systems, volume 33, pages 1877--1901, 2020.

\bibitem{li2023scaling}
Xian Li, Yuxuan Jiang, Yutian Xu, Jingfeng Xue, Yujia Gao, Chao Huang, Yao Qian, Ke Hu, and others.
\textit{Scaling Speech Technology to 1,000+ Languages}.
arXiv preprint arXiv:2305.13516, 2023.

\end{thebibliography}
\end{document}